\documentclass[conference]{IEEEtran}
\IEEEoverridecommandlockouts

\usepackage{cite}
\usepackage{amsmath,amssymb,amsfonts}
\usepackage{graphicx}
\usepackage{textcomp}
\usepackage{xcolor}
\usepackage{tabularx}
\usepackage{booktabs}
\usepackage{url}
\usepackage{orcidlink}

\begin{document}

\title{From Lexical Baselines to Agentic Retrieval-Augmented Generation: Structured Skill and Responsibility-Level Extraction with the SFIA Framework}

\author{
\IEEEauthorblockN{
Ranuga Disansa\IEEEauthorrefmark{1}\,\orcidlink{0009-0005-7140-4584},
U. S. Samarasinghe\IEEEauthorrefmark{2}\,\orcidlink{0009-0007-4932-569X},
and Lasith Gunawardena\IEEEauthorrefmark{2}\,\orcidlink{0000-0002-2495-4103}
}
\IEEEauthorblockA{
\IEEEauthorrefmark{1}\textit{Informatics Institute of Technology},
Colombo, Sri Lanka
}
\IEEEauthorblockA{
\IEEEauthorrefmark{2}\textit{Department of Information Technology},
\textit{University of Sri Jayewardenepura},
Colombo, Sri Lanka
}
}

\maketitle

\begin{abstract}
Automated skill extraction underpins workforce planning, yet most systems represent skills as flat labels with no notion of the responsibility level at which a skill is practiced. The Skills Framework for the Information Age (SFIA) captures exactly this dimension, defining 147 professional skills across seven responsibility levels, but no automated LLM-based extraction targeting SFIA has been reported. We formalize the task as structured prediction of (skill, level) pairs from free text and ask three questions: how accurately can text be mapped onto SFIA's closed vocabulary, which strategies reliably predict the level alongside the skill, and do agentic designs improve on simpler retrieval and prompting? We evaluate five strategies (a lexical baseline, dense retrieval with LLM reranking, a zero-shot schema-constrained LLM, single-agent agentic RAG, and a three-agent retriever--matcher--verifier crew) against expert-mapped European ICT role profiles, all drawing on an SFIA~9 corpus built by a fully automated agentic pipeline that we release. Retrieval-based matching identifies the most skills while generative strategies are markedly more precise; only strategies assigning the level as an explicit decision predict it reliably, with similarity-based selection more than twice as inaccurate; and the crew doubles latency without improving accuracy, so added agent roles do not automatically benefit closed-taxonomy matching. These results provide the first reproducible baseline for structured, level-aware skill extraction against SFIA.
\end{abstract}

\begin{IEEEkeywords}
skill extraction, SFIA, retrieval-augmented generation, agentic AI, large language models, competency frameworks
\end{IEEEkeywords}

\section{Introduction}\label{sec:introduction}

Workforce planning entails having the right people, with the right skills, at the right time and cost \cite{santos2009workforce}. In the IT industry this has grown increasingly complex as technology evolves and roles diversify, so effective planning depends on an accurate, structured understanding of employee skills to support recruitment, allocation, and development decisions.

Hiring and workforce-development data is overwhelmingly free text, so any system that reasons about skills must extract them, normalize them to a shared vocabulary, and compare them across people, roles, and courses. Survey work treats skill extraction as a core NLP task and identifies inconsistent terminology as a persistent obstacle~\cite{senger2024survey}, while supervised approaches such as SkillSpan~\cite{zhang2022skillspan} require annotated data whose value decays as terminology drifts. Large language models (LLMs) change the character of the extraction step, decomposing conjoined and implicit skill mentions in ways strict span metrics penalize~\cite{nguyen2024rethinking}. A label such as ``cloud architecture'', however, says nothing about \emph{level}: whether the person follows instructions, works autonomously, leads delivery, or sets strategy. Systems that represent skills as unnormalized text, including embedding-based applicant-tracking approaches~\cite{resume2vec2025}, do not model this depth dimension at all.

The field has begun normalizing extracted skills onto structured taxonomies, most prominently ESCO~\cite{clavie2023esco,skillmo2025}. We argue for the Skills Framework for the Information Age (SFIA) as the normalization target for professional digital work. Unlike ESCO and O*NET, SFIA defines every skill at one or more of seven generic responsibility levels characterized by autonomy, influence, complexity, and business skills; it targets the global market; and it ships in versioned releases, so adapting a retrieval-based system to a new version reduces to re-indexing rather than re-annotation. SFIA is already used as a mapping target in curriculum design and industry skill mapping~\cite{mapping2016sfia,vonkonsky2016sfia, sfia2024}, but we find no prior automated LLM-based extraction against it. This gap motivates three research questions: (RQ1) how accurately can free text be mapped onto SFIA's closed vocabulary; (RQ2) which strategies reliably predict the responsibility level alongside the skill; and (RQ3) whether agentic and multi-agent designs improve over simpler retrieval and prompting.

This paper makes four contributions:
\begin{itemize}
\item We formalize free-text-to-SFIA matching as structured prediction, returning (skill, level) pairs from the closed SFIA vocabulary, grounded in the 30 expert-mapped European ICT role profiles of CWA 16458:2018, and release all matchers, the corpus pipeline, and the evaluation harness\footnote{\url{https://github.com/prdai/sfia-structured-skill-extraction}}.
\item We describe a fully automated agentic pipeline that turns the public SFIA~9 documentation into a validated 669-record retrieval corpus, with zero out-of-range generations and a residual error of 0.45\% that is dropped.
\item We compare five strategies (BM25, dense retrieval with LLM reranking, zero-shot schema-constrained LLM, single-agent agentic RAG, and a retriever--matcher--verifier crew) on a single corpus and harness, with model sweeps for the three generative strategies.
\item We report a negative result: the crew doubles the single agent's latency without improving F1 and with worse level fidelity, evidence that multi-agent decomposition is not automatically beneficial for closed-taxonomy matching.
\end{itemize}

Section~\ref{sec:related} reviews prior work, Section~\ref{sec:methodology} describes the corpus, evaluation data, and matching strategies, Section~\ref{sec:results} presents the results and threats to validity, and Section~\ref{sec:conclusion} concludes.

\section{Related Work}\label{sec:related}

\subsection{Skill Extraction and Normalization}
Computational job-market analysis treats skill extraction and classification as established NLP tasks~\cite{senger2024survey}. SkillSpan provides annotated spans for supervised extraction~\cite{zhang2022skillspan}, and LLMs trail supervised extractors on strict span metrics while handling conjoined mentions differently~\cite{nguyen2024rethinking}. For normalization, SkiLLMo maps skills onto ESCO with transformers~\cite{skillmo2025}, and Clavi\'e and Souli\'e match free text to ESCO zero-shot with retrieval-shortlisted LLM prompting~\cite{clavie2023esco}. The latter is the closest precedent to our zero-shot and RAG matchers, but it operates over a taxonomy without responsibility levels. Taxonomy-guided occupation classification~\cite{achananuparp2026taxonomy} is the nearest structured-label analogue. None of these target SFIA, and none predict a level alongside the skill.

\subsection{Retrieval-Augmented Generation}
RAG grounds LLM generation in retrieved evidence~\cite{lewis2020rag}, including with frozen models at inference time~\cite{ram2023incontext}; Gao et al.\ survey the design space~\cite{gao2023ragsurvey}. In the retrieval layer, dense retrieval~\cite{karpukhin2020dpr} and sparse lexical scoring with BM25~\cite{robertson2009bm25} capture complementary signals and err on different inputs~\cite{luan2021sparse,thakur2021beir,chen2022spar}, and hybrid combinations improve RAG accuracy~\cite{sawarkar2024blended}. We therefore expose a dense and a sparse retriever to our agents as separate tools, letting the agent decide how to combine evidence rather than applying a fixed rank-fusion step such as RRF~\cite{cormack2009rrf}.

\subsection{Agentic and Multi-Agent Systems}
Agentic RAG couples retrieval with iterative reasoning and tool use~\cite{singh2025agenticrag}, building on ReAct-style interleaving~\cite{yao2023react} and self-reflective retrieval~\cite{asai2024selfrag}. Role-play prompting improves zero-shot reasoning~\cite{kong2024roleplay}, which motivates the role/goal/backstory structure of CrewAI~\cite{moura2023crewai}. Multi-agent systems decompose work across role-specialized agents~\cite{wu2024autogen,hong2024metagpt,qian2024chatdev}; multi-agent debate improves the factuality of generated output~\cite{du2024debate} and multi-agent judging improves evaluation quality~\cite{chan2024chateval}, both of which motivate the verifier agents in our pipeline and matcher. Multi-agent resume screening exists~\cite{screening2025multiagent}, but its agents exchange ad-hoc structured components rather than labels from a closed competency taxonomy; we find no prior multi-agent work over one.

\section{Methodology}\label{sec:methodology}

\subsection{Task Definition}
Given free text $t$ (a job description, course outline, or profile summary), a matcher returns a set of pairs $\{(s_i, \ell_i)\}$, where $s_i$ is one of the 147 SFIA skill names and $\ell_i \in \{1,\dots,7\}$ is a level defined for that skill.

\subsection{SFIA Corpus Construction}\label{sec:corpus}
\begin{figure*}
\centering
\includegraphics[width=0.6\textwidth]{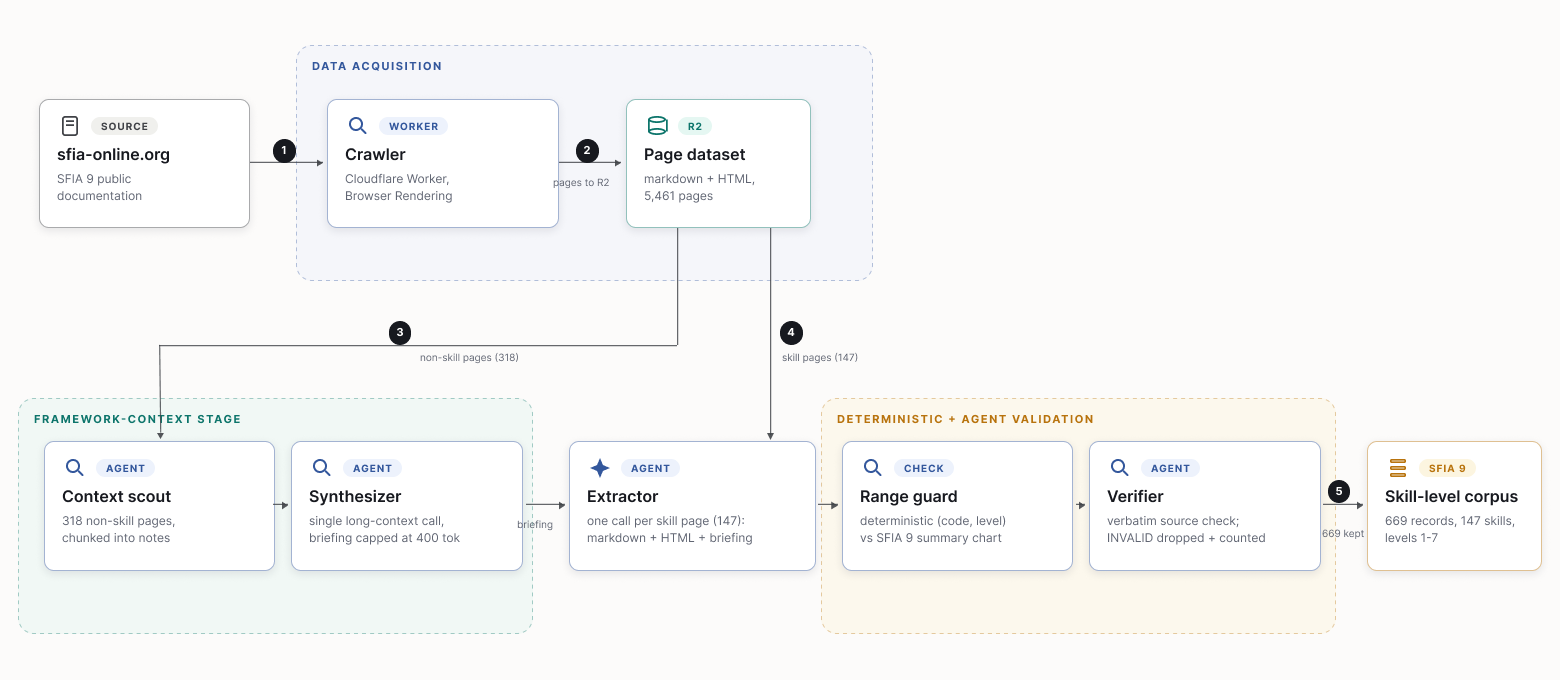}
\caption{Fully automated corpus-construction pipeline: a serverless crawler stores each documentation page as markdown and HTML; a context-scouting agent reads the non-skill pages and a synthesizer agent condenses its notes into a standing briefing; an extractor agent emits (code, level, text) records per skill page; a deterministic guard checks every pair against the SFIA~9 skill--level chart; a verifier agent enforces verbatim fidelity, with rejections dropped and counted.}
\label{fig:arch-extraction}
\end{figure*}

The corpus consists of \emph{skill-level records}, each pairing one SFIA skill (identified by its four-letter code) with the verbatim description of that skill at one level. It is built from the public SFIA~9 documentation by the fully automated pipeline of Fig.~\ref{fig:arch-extraction}; no manual correction is applied at any stage.

\subsubsection{Acquisition} A Cloudflare Worker crawls \mbox{sfia-online.org} through a managed browser-rendering service, storing each of the 5,461 reachable pages as markdown and raw HTML. Downstream stages use the 147 per-skill pages and the 318 further documentation pages with extractable content; the remainder is unused. Using an LLM instead of hand-written per-source parsers follows work on structured views of heterogeneous documents~\cite{evaporate2023}: the markup supplies structural cues and the linearized text serves as the copy source.

\subsubsection{Framework-context stage} A scout agent reads the 318 non-skill pages in fixed-size chunks and emits notes on framework-wide conventions (114 pages yielded relevant notes). A synthesizer agent then condenses all notes in one long-context call into a standing briefing, hard-capped at 400 output tokens and prepended to every extractor prompt as background only, never as a source of extracted text. This design follows prior work on gist memories~\cite{readagent2024}, ahead-of-use corpus summarization~\cite{graphrag2024,raptor2024}, contextual retrieval~\cite{merola2025contextual}, and the finding that a single long-context pass is competitive with retrieval-based decomposition when the input fits the window~\cite{li2024selfroute}.

\subsubsection{Extraction} One agent call per skill page (147 pages) receives the page markdown, the HTML trimmed to its main content block, and the briefing, and returns structured records \{code, level, text, reasoning\} under a JSON schema. The prompt states a single structural criterion (a skill-named level header opens a record) and never enumerates which levels exist, so the page content alone determines which (code, level) pairs are produced.

\subsubsection{Validation} A \emph{deterministic range guard} rejects any (code, level) pair that the official SFIA~9 skill--level chart does not define. The chart is the pipeline's one manually curated reference artifact, transcribed once into JSON and cross-checked against all 147 skill pages; ``fully automated'' thus describes the per-run extraction given this fixed reference. A \emph{verifier agent} then re-reads the source page and rejects any record that is not a verbatim, complete copy of its section, in the spirit of multi-agent debate and LLM judging~\cite{du2024debate,chan2024chateval}. Rejections are dropped and counted.

\subsubsection{Outcome} The extractor produced exactly the 672 (code, level) pairs the chart defines, so the range guard fired zero times, and the verifier rejected 3 records (0.45\%). The released corpus holds 669 records covering all 147 skills.

\subsection{Evaluation Data and Metrics}
Ground truth is the 30 European ICT professional role profiles of CWA 16458:2018, each with a one-sentence summary statement and an expert-assigned set of SFIA (skill, level) mappings (176 pairs over 74 distinct skills). The profiles predate SFIA~9: 20 of the 74 gold skill names (54 of 176 pairs, 31\%) were renamed or retired before SFIA~9 and cannot be matched verbatim by any system over the SFIA~9 vocabulary, which caps achievable recall at 0.69 for every strategy. All 122 remaining pairs carry levels that SFIA~9 defines for their skill. Each matcher receives only the summary statement and must return all applicable pairs, one query per role. Metrics are macro-averaged over the 30 roles: per-role set precision, recall, and F1 over predicted skill names, and, over each role's correctly named skills, exact and within-one level accuracy and level mean absolute error (MAE)~\cite{baccianella2009ordinal}. Predictions are deduplicated, a failed call scores as an empty prediction for every strategy, and mean wall-clock seconds per record is reported. The task is thus closed-vocabulary multi-label classification with ordinal level prediction.

\subsection{Matching Strategies}
\begin{figure*}
\centering
\begin{minipage}[t]{0.49\textwidth}
\centering
\includegraphics[width=\linewidth]{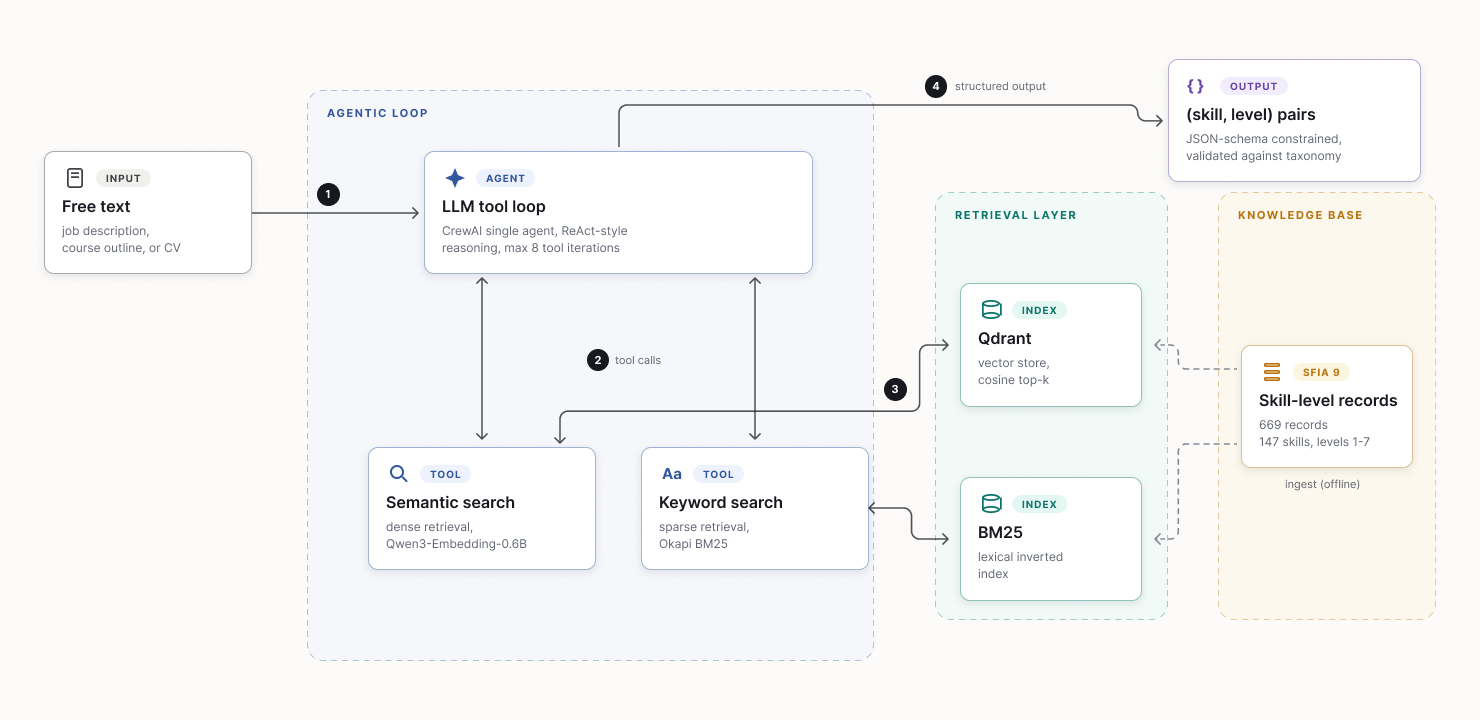}
\caption{Single-agent agentic RAG: one LLM tool loop with dense (Qdrant) and sparse (BM25) retrieval tools over the SFIA~9 skill-level corpus, emitting schema-validated (skill, level) pairs.}
\label{fig:arch-single}
\end{minipage}\hfill
\begin{minipage}[t]{0.49\textwidth}
\centering
\includegraphics[width=\linewidth]{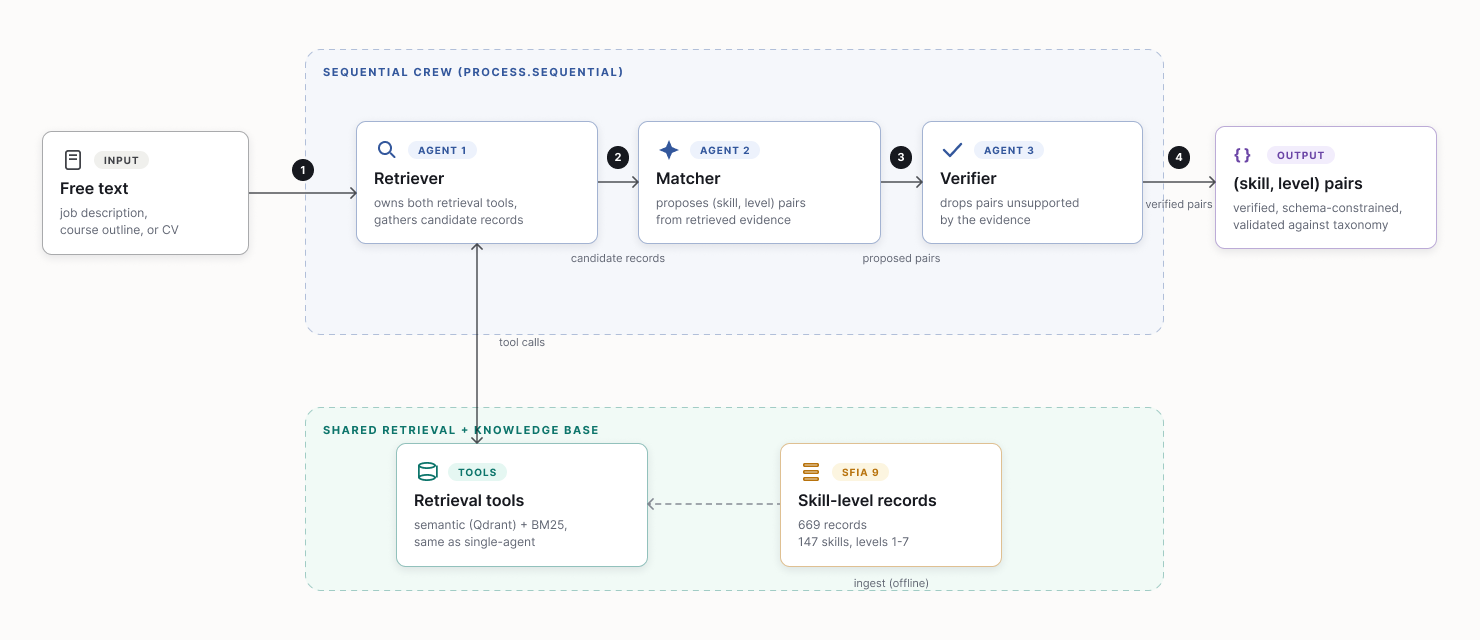}
\caption{Multi-agent crew: sequential retriever--matcher--verifier pipeline sharing the same retrieval tools and corpus as the single-agent matcher.}
\label{fig:arch-crew}
\end{minipage}
\end{figure*}

All five matchers draw on the same corpus (as retrieval substrate or, for the zero-shot matcher, as its closed output vocabulary) and are scored by the same harness. Each otherwise keeps its own idiomatic decision rules, stated below, so the comparison is a snapshot of representative approaches rather than a knob-matched head-to-head. Fig.~\ref{fig:arch-single} shows the single-agent architecture and Fig.~\ref{fig:arch-crew} the crew.

\subsubsection{Lexical baseline (BM25)} Okapi BM25~\cite{robertson2009bm25} with default $k_1$ and $b$ over the skill-level records; every record scoring at least 0.7 of the query's top score is returned, a relative threshold rather than a fixed $k$.

\subsubsection{Dense retrieval with LLM reranking} Each record is embedded (Qwen3-Embedding-0.6B) as ``skill name: level text'' into a Qdrant vector store. The query retrieves the top 20 by cosine similarity, each candidate is scored pointwise for relevance by Llama-3.3-70B, and candidates scoring $\geq 0.6$ are returned.

\subsubsection{Zero-shot LLM} A single LLM call with no retrieval: the model maps text to SFIA pairs from parametric knowledge alone, with output forced valid by a JSON schema whose skill field enumerates the 147 skill names and whose level is an integer 1--7.

\subsubsection{Single-agent agentic RAG} A CrewAI agent~\cite{moura2023crewai} holding two retrieval tools, semantic search over the Qdrant index and BM25 keyword search, called iteratively (up to 8 iterations) before emitting schema-validated (skill, level) pairs, post-validated against the canonical vocabulary. Serving is a serverless inference catalog (Cloudflare Workers AI) behind an OpenAI-compatible gateway.

\subsubsection{Multi-agent crew} Three sequential CrewAI agents sharing one model: a \emph{retriever} owning both tools, a \emph{matcher} proposing pairs from the retrieved evidence, and a \emph{verifier} that discards pairs unsupported by the evidence, following the tradition of agents judging other agents' output~\cite{du2024debate,chan2024chateval}. Task outputs chain as context.

\subsection{Model Selection}
Function-calling ability varies widely across models~\cite{patil2025bfcl} and no external benchmark covers this serving stack, so models were selected empirically with one procedure for the three generative strategies: sweep each candidate over the full 30-role harness at temperature 0 and adopt the best-F1 model. Pools differ per strategy, and the selection shares the evaluation set, a limitation discussed in Section~\ref{sec:threats}. The zero-shot matcher swept 10 models (Table~\ref{tab:sweep-llm}, winner Gemma-4-26B); the single agent swept 8 function-calling models (Table~\ref{tab:sweep-single}, winner GLM-4.7-Flash); the crew swept 5 (Table~\ref{tab:sweep-crew}, winner Kimi-K2.6; two further models produced malformed tool calls in multi-turn conversation and were excluded). All sweeps and final runs ran strictly serially to avoid rate-limit interference.

\section{Results and Discussion}\label{sec:results}

\begin{table}
\centering
\caption{Best configuration per strategy on the 30-role harness. Each row is a single reporting run of the model its sweep selected; the sweeps (Tables~\ref{tab:sweep-llm}--\ref{tab:sweep-crew}) are separate selection runs.}
\label{tab:main-results}
\footnotesize
\setlength{\tabcolsep}{3.5pt}
\begin{tabularx}{\linewidth}{Xrrrrrrr}
\toprule
\textbf{Strategy} & \textbf{P} & \textbf{R} & \textbf{F1} & \textbf{MAE} & \textbf{Exact} & \textbf{W1} & \textbf{s/rec}\\
\midrule
BM25 lexical & 0.141 & 0.133 & 0.128 & 0.94 & 0.260 & 0.875 & 0.001\\
Dense + rerank & 0.266 & 0.243 & \textbf{0.232} & 1.60 & 0.181 & 0.436 & 6.17\\
Zero-shot LLM & 0.436 & 0.117 & 0.176 & 0.74 & 0.368 & \textbf{0.895} & 0.23\textsuperscript{a}\\
Agentic RAG & 0.361 & 0.172 & 0.213 & \textbf{0.68} & \textbf{0.526} & 0.816 & 167.1\\
Multi-agent crew & 0.398 & 0.218 & 0.212 & 1.16 & 0.319 & 0.717 & 327.1\\
\bottomrule
\multicolumn{8}{p{0.95\linewidth}}{\footnotesize Macro-averages over the 30 roles; F1 is the mean of per-role F1, not the harmonic mean of the P and R columns. MAE/Exact/W1 = level MAE, exact and within-one level accuracy on correctly named skills; s/rec = mean wall-clock seconds per record. \textsuperscript{a}From the sweep run; the final run repeated identical temperature-0 prompts and was largely served from the gateway response cache (0.05 s/record).}\\
\end{tabularx}
\end{table}

Table~\ref{tab:main-results} reports each strategy's best swept configuration; three findings stand out.

\textbf{Dense retrieval attains the highest observed F1, and LLM strategies trade recall for precision.} The dense pipeline records the highest skill F1 (0.232) via the highest recall (0.243), though its margin over the agentic strategies (0.213, 0.212) is of the same order as run-to-run variance, so we claim no strict ranking. The robust pattern is the trade-off structure: the zero-shot matcher is the precision extreme (0.436 at recall 0.117), and retrieval tools buy recall back (0.172 single-agent, 0.218 crew) while keeping precision well above the dense pipeline's 0.266. Which profile is preferable depends on the application: recall matters for a human reviewer, precision for an automated pipeline. Recall is capped at 0.69 for every strategy by the gold standard's pre-SFIA-9 vocabulary.

\textbf{Level assignment requires a generative decision, but not necessarily an agent.} On the level dimension, precisely what distinguishes SFIA from flat taxonomies, the ranking inverts. Dense retrieval, despite the highest F1, is the worst level predictor by far (MAE 1.60, within-one 0.436): its level is a by-product of which record embeds closest, and even its 70B reranker, which sees the level text, scores relevance rather than committing to a level. Strategies in which an LLM assigns the level as an explicit output sit in a different regime: the single agent deliberating over retrieved descriptors (MAE 0.68, exact 0.526 final; 0.95 in the sweep run of the same configuration) and the schema-constrained zero-shot LLM (MAE 0.74, within-one 0.895), as does the lexical baseline, whose top record carries its level along (0.94). This is deliberately a group-level contrast: the gap to dense retrieval far exceeds run-to-run variance, while the ordering within the group does not.

\textbf{More agents is not better.} The crew matches the single agent on skill F1 (0.212 versus 0.213) at double the latency (327 versus 167 s/record, since three sequential agents each make their own LLM round trips), with markedly worse level fidelity (MAE 1.16 versus 0.68). Passing evidence as conversation text appears to lose the descriptor grounding the single agent gets from reading tool output directly. The verifier raises precision (0.398 versus 0.361) and recall (0.218 versus 0.172) slightly, but nothing justifies the cost, consistent with multi-agent decomposition helping open-ended generation more than closed-label prediction.

\begin{table}
\centering
\caption{Zero-shot LLM model sweep (30 roles).}
\label{tab:sweep-llm}
\footnotesize
\setlength{\tabcolsep}{3.5pt}
\begin{tabularx}{\linewidth}{Xrrrr}
\toprule
\textbf{Model} & \textbf{F1} & \textbf{Lvl exact} & \textbf{Lvl MAE} & \textbf{failed}\\
\midrule
Gemma-4-26B-A4B & \textbf{0.176} & 0.368 & 0.74 & 0\\
Nemotron-3-120B-A12B & 0.174 & 0.438 & 0.88 & 0\\
Llama-3.1-8B-fast & 0.166 & 0.194 & 1.64 & 0\\
Llama-4-Scout-17B & 0.153 & \textbf{0.694} & \textbf{0.36} & 0\\
Qwen3-30B-A3B & 0.152 & 0.400 & 0.93 & 0\\
Mistral-Small-3.1-24B & 0.139 & 0.393 & 0.75 & 0\\
Kimi-K2.6 & 0.136 & 0.286 & 0.79 & 3\\
Kimi-K2.7-Code & 0.134 & 0.500 & 0.50 & 5\\
GLM-5.2 & 0.123 & 0.500 & 0.58 & 7\\
GLM-4.7-Flash & 0.100 & 0.600 & 0.40 & 9\\
\bottomrule
\multicolumn{5}{p{0.95\linewidth}}{\footnotesize Failed roles: the model exhausted its output budget on hidden reasoning under schema-constrained decoding and returned no content; such roles score as empty predictions.}\\
\end{tabularx}
\end{table}

Table~\ref{tab:sweep-llm} gives the zero-shot sweep. Skill F1 and level accuracy rank models very differently. Llama-4-Scout is mid-pack on F1 yet posts the best level numbers (exact 0.694, MAE 0.36, on the small set it names correctly), so level-critical deployments would weigh models differently than coverage-critical ones. Four of the ten models intermittently spent their whole output budget on hidden reasoning under schema-constrained decoding and returned nothing, so robustness to constrained decoding proved as decisive as capability.

\begin{table}
\centering
\caption{Single-agent agentic RAG model sweep (30 roles).}
\label{tab:sweep-single}
\footnotesize
\setlength{\tabcolsep}{3.5pt}
\begin{tabularx}{\linewidth}{Xrrrrr}
\toprule
\textbf{Model} & \textbf{F1} & \textbf{Lvl exact} & \textbf{Lvl MAE} & \textbf{s/rec} & \textbf{failed}\\
\midrule
GLM-4.7-Flash & \textbf{0.236} & 0.37 & 0.95 & 227 & 1\\
Mistral-Small-3.1-24B & 0.224 & 0.25 & 0.95 & 14 & 1\\
Kimi-K2.6 & 0.216 & 0.52 & 0.67 & 151 & 0\\
Nemotron-3-120B-A12B & 0.188 & \textbf{0.53} & \textbf{0.65} & 155 & 0\\
Gemma-4-26B-A4B & 0.176 & 0.38 & 0.84 & 244 & 4\\
GLM-5.2 & 0.086 & 0.50 & 0.50 & 20 & 22\\
Kimi-K2.7-Code & 0.060 & 0.67 & 0.33 & 25 & 21\\
Llama-4-Scout-17B\textsuperscript{a} & 0.000 & --- & --- & 4 & 0\\
\bottomrule
\multicolumn{6}{l}{\footnotesize\textsuperscript{a}Never invoked the retrieval tools despite catalog capability flag.}\\
\multicolumn{6}{l}{\footnotesize Failed roles score as empty predictions (see Table~\ref{tab:sweep-llm} note).}\\
\end{tabularx}
\end{table}

\begin{table}
\centering
\caption{Multi-agent crew model sweep (30 roles, one shared model).}
\label{tab:sweep-crew}
\footnotesize
\setlength{\tabcolsep}{3.5pt}
\begin{tabularx}{\linewidth}{Xrrrrr}
\toprule
\textbf{Model} & \textbf{F1} & \textbf{Lvl exact} & \textbf{Lvl MAE} & \textbf{s/rec} & \textbf{failed}\\
\midrule
Kimi-K2.6 & \textbf{0.189} & 0.29 & 1.20 & 342 & 0\\
Nemotron-3-120B-A12B & 0.172 & 0.32 & \textbf{0.87} & 401 & 3\\
GLM-4.7-Flash & 0.160 & \textbf{0.34} & 0.92 & 315 & 1\\
GLM-5.2 & 0.073 & 0.50 & 0.50 & 44 & 24\\
Kimi-K2.7-Code & 0.059 & 0.40 & 0.60 & 78 & 21\\
\bottomrule
\multicolumn{6}{l}{\footnotesize Failed roles score as empty predictions (see Table~\ref{tab:sweep-llm} note).}\\
\end{tabularx}
\end{table}

Tables~\ref{tab:sweep-single} and~\ref{tab:sweep-crew} give the agentic sweeps. The single-agent ranking does not carry over: the single-agent winner (GLM-4.7-Flash) drops to third in the crew, while Kimi-K2.6, third as a single agent, wins the crew. Multi-turn multi-agent robustness is therefore not predicted by single-agent tool use, consistent with the cross-setting variance that function-calling benchmarks report~\cite{patil2025bfcl}. Reliability, not capability, separates the bottom rows: Kimi-K2.7-Code and GLM-5.2 fail most roles in both configurations (breakdown in long tool conversations and reasoning-budget exhaustion), and Llama-4-Scout silently ignores its tools.

\subsection{Threats to Validity}\label{sec:threats}
The harness contains 30 roles with one-sentence summaries, so results may differ on longer inputs. Ground-truth mappings predate SFIA~9 (Section~\ref{sec:corpus}: 31\% of gold pairs unreachable, recall capped at 0.69). Model selection used the same 30 roles as the final evaluation; with no held-out split, per-strategy winners, and small sweep margins are configuration-level observations, not generalization claims. Each configuration is evaluated in a single run, and identical configurations moved between runs by more than several reported margins (crew: F1 0.189 sweep versus 0.212 final; single agent: F1 0.236 and MAE 0.95 sweep versus 0.213 and 0.68 final), so we interpret only differences that far exceed this variance and flag near-ties as such. Level metrics condition on each strategy's own correctly named skills, so denominators differ; a joint (skill, level) metric is left for future work. Zero-shot latencies are partially cache-deflated, all serving ran on one provider's catalog, and function-calling behavior varies across settings~\cite{patil2025bfcl}, so rankings may not transfer to other stacks. Finally, the corpus excludes 3 of 672 chart-defined records due to verifier rejections (at least two spurious on inspection), reported rather than repaired to keep the pipeline fully automated.

\section{Conclusion}\label{sec:conclusion}

This work began from a practical limitation in automated skill extraction, identifying a skill is not sufficient when workforce decisions also depend on the level of responsibility at which that skill is exercised. By treating the task as a structured prediction of SFIA (skill, level) pairs, we show that this additional dimension can be extracted automatically from free text within a closed and versioned professional framework. To our knowledge, this is the first systematic evaluation of LLM-based extraction targeting SFIA, supported by a fully automated corpus-construction pipeline whose output is explicitly validated and whose residual error is characterized.

The experiments also clarify what an effective solution to this problem should look like. Retrieval is valuable for identifying candidate skills and improving coverage, while responsibility-level prediction is substantially more reliable when the model is required to make the level an explicit decision rather than inheriting it from the nearest retrieved record. At the same time, increasing architectural complexity does not inherently improve the result: the three-agent crew roughly doubles the latency of the single-agent approach without producing a corresponding improvement in skill matching or level fidelity. Taken together, these findings point toward a simpler hybrid design, retrieval for candidate discovery followed by explicit model-based skill and level adjudication, rather than progressively deeper agent decomposition.

The result is therefore more than a comparison of five matching strategies, it establishes a reproducible path from unstructured descriptions to normalized, level-aware competency representations. Instead of leaving job descriptions, professional profiles, and course descriptions as incompatible collections of free-text skills, they can be mapped into the same SFIA vocabulary and compared at both the skill and responsibility levels. This provides a foundation for downstream workforce-planning systems in which requirements, capabilities, and development needs can be represented using the same structured language.

Future work should test this approach on larger and more current SFIA-aligned datasets, characterize agentic variance through repeated runs, and evaluate the matcher as the normalization layer of end-to-end applications such as applicant tracking and workforce gap analysis. A particularly promising direction is a hybrid architecture that combines the dense retriever's stronger coverage with the explicit level-assignment behavior of the generative approaches, directly building on the complementary strengths identified in this study.

\bibliographystyle{IEEEtran}
{\footnotesize
\bibliography{literature}}

\end{document}